\documentclass{article}

\usepackage[numbers,sort&compress]{natbib}

 \usepackage[eandd, final]{neurips_2026}

\usepackage[utf8]{inputenc} 
\usepackage[T1]{fontenc}    
\usepackage{hyperref}       
\usepackage{url}            
\usepackage{booktabs}       
\usepackage{amsfonts}       
\usepackage{nicefrac}       
\usepackage{microtype}      
\usepackage{xcolor}         

\usepackage{float} 
\usepackage{stfloats} 

\usepackage{url}
\usepackage{tabularray}
\UseTblrLibrary{booktabs}
\usepackage{graphicx}
\usepackage{multirow}
\usepackage{bm}          

\usepackage{amsmath}
\usepackage{multirow}
\usepackage{colortbl}
\renewcommand\arraystretch{0.95}
\usepackage{wrapfig}
\usepackage{xspace}
\usepackage{arydshln}
\usepackage[font=small,labelfont=bf]{caption}
\usepackage{subcaption}
\usepackage{paralist}
\usepackage{enumitem}
\usepackage{pifont}
\usepackage{hyperref}
\definecolor{linkblue}{HTML}{3C83E6}
\hypersetup{
    colorlinks=true,
    linkcolor=linkblue,   
    citecolor=linkblue,   
    filecolor=magenta,      
    urlcolor=magenta
}

\usepackage{ulem}
\usepackage{cleveref}
\usepackage[most]{tcolorbox}
\usepackage{amssymb}

\definecolor{darksalmon}{rgb}{0.91, 0.59, 0.48}
\definecolor{gainred}{HTML}{E45D35}  
\definecolor{gainblue}{HTML}{4C7CD4}  

\usepackage{makecell}
\usepackage{bbding}

\definecolor{forestgrhl}{RGB}{84, 186, 111}
\definecolor{darkgrhl}{RGB}{136, 219, 158}
\definecolor{greenhl}{RGB}{201, 242, 212}
\definecolor{faintgrhl}{RGB}{244, 255, 241}
\definecolor{darkredhl}{RGB}{232, 173, 161}
\definecolor{redhl}{RGB}{242, 208, 201}
\definecolor{faintredhl}{RGB}{255, 244, 241}
\definecolor{whitehl}{RGB}{255, 255, 255}
\definecolor{yellowhl}{RGB}{255, 217, 164}
\definecolor{forestgreen}{rgb}{0.13, 0.55, 0.13}
\definecolor{bluehl}{RGB}{201, 225, 242}

\title{A Large-Scale Dataset and Benchmark: Do Protein–Ligand Models Learn Binding Sites or\\ Just Binding Likelihood?}

\author{%
  \textbf{Zhaohan Meng}\textsuperscript{1}$,
  \textbf{Zhen Bai}\textsuperscript{2}$,
  \textbf{Ke Yuan}\textsuperscript{3,4},
  \textbf{Iadh Ounis}\textsuperscript{1},
  \textbf{Zaiqiao Meng}\textsuperscript{1,5},
  \textbf{Hao Xu}\textsuperscript{6},
  \textbf{Joseph Loscalzo}\textsuperscript{6,7} \\[2mm]
  \textsuperscript{1}School of Computing Science, \textsuperscript{3}School of Cancer Sciences, University of Glasgow\\
  \textsuperscript{2}School of Life Science and Technology, Institute of Science Tokyo\\
  \textsuperscript{4}Cancer Research UK Scotland Institute\\
  \textsuperscript{5}Language Technology Lab, University of Cambridge\\
  \textsuperscript{6}Department of Medicine, Brigham and Women's Hospital, Harvard Medical School\\
  \textsuperscript{7}The Broad Institute of MIT and Harvard \vspace{1mm}\\
\parbox{0.03\textwidth}{\includegraphics[width=\linewidth]{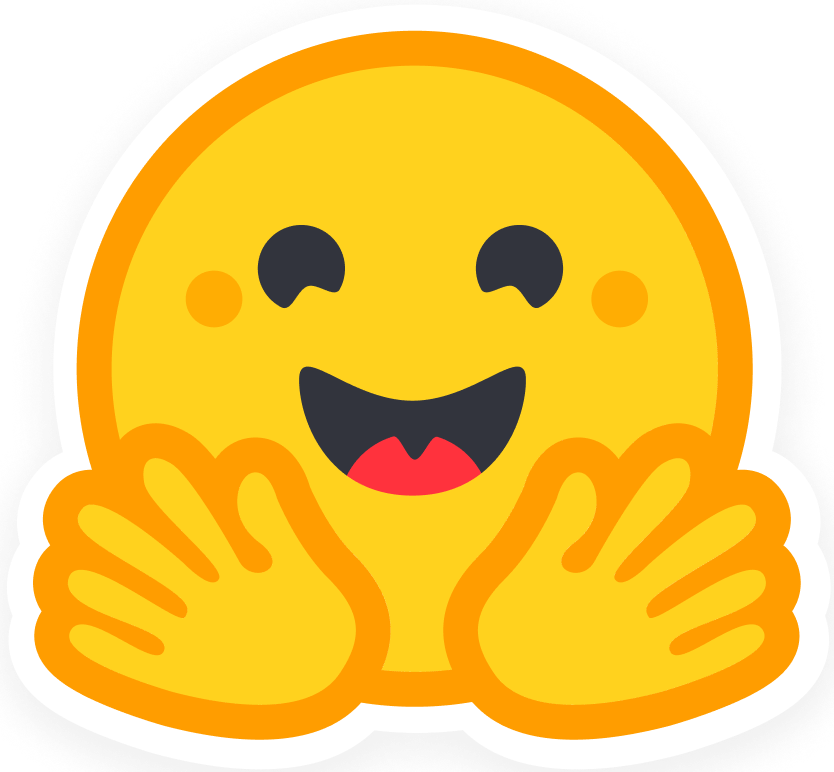}}\hspace{1mm}\href{https://huggingface.co/datasets/Zhaohan-Meng/InteractBind}{\texttt{https://huggingface.co/datasets/Zhaohan-Meng/InteractBind}}\vspace{1mm}\\
\texttt{z.meng.3@research.gla.ac.uk}; \hspace{3mm} \texttt{haxu@bwh.harvard.edu}\vspace{-3mm}
}

\begin{document}

\maketitle

\begin{abstract}
Protein--ligand modeling underpins computational drug discovery and molecular design. Existing protein--ligand benchmarks typically evaluate whether a protein and ligand interact and how strongly they bind, through tasks such as binary binding prediction and affinity regression. However, these evaluations provide limited evidence of whether models can localize binding sites or identify the non-covalent interactions underlying molecular recognition. To address this gap, we introduce \textbf{InteractBind}, a large-scale protein--ligand dataset comprising approximately 100k protein--ligand pairs, together with a benchmark for fine-grained evaluation. The core fine-grained task is that of binding-site localization, which uses protein-residue and ligand-atom interaction maps spanning six major types of non-covalent interactions to assess whether model-derived interaction maps localize binding sites. InteractBind further includes binding affinity and protein similarity-controlled splits to support realistic generalization assessment. Using InteractBind, we evaluate eight existing sequence-based and interaction-aware models, assessing binary binding prediction and binding-site localization. Results reveal limited binding-site localization despite strong binary binding prediction, with marked variation across non-covalent interaction types. Overall, InteractBind establishes a benchmark paradigm that encourages the development of more interpretable and physically grounded protein--ligand models.
\end{abstract}
\vspace{-0.5em}
\section{Introduction}

Protein--ligand interactions underlie a wide range of biological and chemical processes, including enzymatic catalysis, molecular recognition, cellular signaling, and therapeutic modulation~\cite{barabasi2004network,tan2024molecular,schwartz2009enzymatic}. Experimentally characterizing these interactions at scale remains costly and time-consuming, which makes protein--ligand modeling a central component of computational drug discovery: virtual screening prioritizes candidate binders from large compound libraries, while binding affinity prediction supports lead optimization before wet-lab validation~\cite{finan2017druggable, dara2022machine, wohlwend2025boltz}. Representative advances include structure-based approaches that model three-dimensional binding geometry, such as docking, molecular dynamics, and geometric deep learning~\cite{adams2025shepherd,morehead2025assessing,minan2025informed}, as well as sequence-based approaches that predict binding from protein sequences and ligand string or graph representations~\cite{chen2025semi,zhang2025labind,jia2026deep}. To evaluate these approaches, there are existing benchmarks for protein--ligand modeling focusing primarily on the binary binding prediction~\cite{gilson2016bindingdb,zitnik2018biosnap} and the binding affinity prediction~\cite{liu2015improving,davis2011comprehensive} tasks. These tasks measure whether a model predicts binding occurrence or binding strength, rather than identifying specific binding sites that explain molecular recognition~\cite{mattos1996locating, vcerny2007non}. Consequently, evidence that existing models capture such fine-grained interaction patterns often comes from a few visualized case studies, motivating a new benchmark that evaluates these capabilities systematically against large-scale ground-truth annotations.

Beyond binary binding and binding affinity prediction, fine-grained evaluation of protein--ligand models should test whether models localize the binding sites that underlie molecular recognition. In experimentally resolved protein--ligand complex structures, binding is mediated by non-covalent contacts between ligand atoms and protein residues (Sec.~\ref{sec:preliminary})~\cite{finan2017druggable,peluso2022recognition}. These contacts naturally define binding sites: a residue is considered a binding-site residue if it participates in one or more ligand-stabilizing contacts~\cite{mattos1996locating,burley2017protein}. Thus, binding-site localization evaluates whether model-derived interaction maps identify where the ligand binds. However, evaluating only the overall binding-site map cannot reveal which physicochemical contact patterns are captured, because binding-site residues arise from interaction types with different geometric and chemical constraints. Retaining interaction-type annotations further enables the evaluation of whether models actually identify the non-covalent interactions that stabilize residue--atom contacts. Such information supports targetable residue identification, selectivity analysis, and structure-informed lead optimization~\cite{crunkhorn2024predicting,holehouse2024molecular}. Existing sequence-based protein--ligand binding benchmarks rarely provide binding-site and interaction-type-specific annotations at scale, limiting systematic evaluation of whether models localize binding sites and identify the non-covalent interaction patterns underlying ligand recognition.

\begin{figure*}[t]
    \centering
    \includegraphics[width=\textwidth]{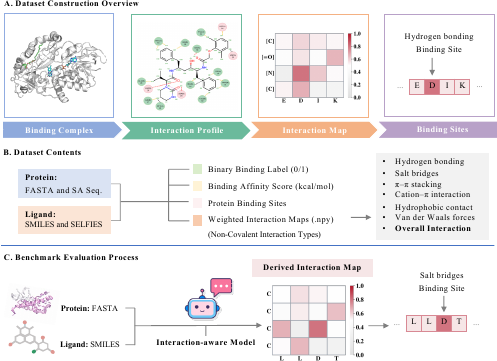}
    \caption{
    \textbf{Overview of InteractBind.}
    \textbf{A.}
    InteractBind is constructed from experimentally resolved protein--ligand complexes by deriving interaction profiles, sequence-level interaction maps, and protein binding-site annotations.
    \textbf{B.}
     Each sample includes protein and ligand sequence representations, binary binding labels, binding affinity values, binding-site annotations, and weighted interaction maps for six non-covalent interactions together with an overall interaction map.
    \textbf{C.}
    Given protein and ligand sequence representations, model-derived interaction maps are evaluated against ground-truth protein binding-site annotations.
    }
    \vspace{-1em}
    \label{fig:benchmark_overview}
\end{figure*}

To address this limitation, we introduce \textbf{InteractBind}, a new large-scale protein--ligand dataset derived from the Protein Data Bank (PDB)~\citep{berman2007worldwide}, and establish the corresponding \textbf{InteractBind benchmark} for the fine-grained evaluation of protein--ligand modeling. The dataset comprises approximately 100k protein--ligand pairs, spanning around 11k proteins and 9k ligands. Existing benchmarks typically provide only binary labels or affinity scores: for example, BioSNAP~\cite{zitnik2018biosnap} provides binary binding labels, while KIBA~\cite{tang2014making} provides scalar binding affinity values. In contrast, as shown in Fig.~\ref{fig:benchmark_overview}, InteractBind also provides sequence-level interaction maps for six major non-covalent interactions, together with an overall interaction map that defines protein binding sites. The new benchmark centers on binding-site localization as the primary fine-grained evaluation task, assessing whether model-derived interaction maps can localize the protein residues involved in ligand binding. The interaction-type-specific maps further support evaluation of whether models identify the non-covalent interactions that stabilize residue--atom contacts. 

To improve benchmark realism and generalization assessment, InteractBind provides two types of dataset splits: an affinity-aware in-distribution (ID) split, and protein similarity-controlled out-of-distribution (OOD) splits for assessing generalization to less familiar protein targets. In the ID split, we benchmark eight existing representative interaction-aware models across binary binding prediction, binding-site localization, and interaction-type-specific evaluation. The best-performing model achieves 98.3\% AUROC for binary binding prediction but only 21.6\% BRHR@1 for binding-site localization, where BRHR@1 measures whether the top-ranked predicted residue matches a ground-truth binding-site residue. Interaction-type-specific evaluation further reveals uneven performance across non-covalent interaction categories, indicating that accurate binding prediction does not imply a reliable identification of the physicochemical contacts that stabilize binding. For OOD evaluation, we construct four protein similarity-controlled dataset splits using global sequence alignment~\cite{hamamsy2024protein}, with maximum train--test protein similarities of 25\%, 28\%, 31\%, and 33\%; we observe that performance decreases under these splits, highlighting the difficulty of localizing binding sites for less familiar protein targets. To our knowledge, no prior sequence-based benchmark provides explicit residue-level interaction supervision together with a systematic evaluation of binding-site localization derived from non-covalent contacts. Our contributions are summarized as follows:
\vspace{-0.5em}
\begin{itemize}
    \item[\ding{182}] We introduce \textbf{InteractBind}, a new large-scale protein--ligand dataset comprising approximately 100k pairs, with binary labels, affinity values, binding-site annotations, and non-covalent interaction maps.
    \vspace{-0.5em}
    \item[\ding{183}] We formulate binding-site localization as the core fine-grained evaluation task, supported by metrics for residue-level localization and interaction-type-specific contact analysis, and construct both ID and OOD splits for realistic generalization.
    \vspace{-0.5em}
    \item[\ding{184}] We benchmark eight representative protein--ligand models and show that interaction-aware architectures provide measurable binding-site localization signals, while highlighting substantial room for improving localization and non-covalent interaction identification.
\end{itemize}

\subsection{Protein--ligand modeling}

Protein--ligand modeling has been studied from both structure-based and sequence-based perspectives. Structure-based methods~\cite{adams2025shepherd,morehead2025assessing,minan2025informed} model binding from 3D structures, including molecular dynamics, geometric deep learning, and recent biomolecular structure prediction models such as AF3~\cite{abramson2024accurate} and Boltz-1~\cite{wohlwend2025boltz}. However, these methods depend on high-quality structures or reliable predicted poses, and geometric accuracy does not necessarily ensure chemically faithful non-covalent interactions~\cite{abramson2024accurate,masters2025investigating}. Sequence-based models provide a scalable alternative when complex structures are unavailable, predicting binding from protein sequences and ligand string or graph representations~\cite{huang2021moltrans,chen2020transformercpi}. Recent interaction-aware models further use attention or fusion modules to model residue--atom correspondences, with qualitative case studies suggesting that learned interaction maps can highlight plausible binding-site residues~\cite{nguyen2023perceiver,zeng2024cat,meng-etal-2025-fusiondti,bai2023interpretable}. However, because existing datasets rarely provide residue--atom interaction annotations, it remains difficult to determine whether these maps systematically localize physically meaningful binding sites or merely correlate with binary binding prediction. Our new InteractBind benchmark is motivated by this gap, providing interaction annotations for evaluating whether sequence-based models localize binding residues and interaction patterns.

\subsection{Protein--ligand datasets}

Protein--ligand datasets serve different purposes depending on the level of information they provide. Structural resources such as the PDB and AlphaFoldDB provide experimentally resolved or predicted protein structures, supporting structural analysis, binding-pose modeling, and structure-based hypothesis generation~\citep{berman2007worldwide,varadi2022alphafold}. Docking-oriented datasets, such as DUD-E, are widely used to evaluate virtual screening methods by providing active compounds and property-matched decoys for target proteins~\cite{mysinger2012directory}. Pair-level learning datasets support complementary predictive tasks: BindingDB, BioSNAP, and Human are commonly used for binary drug--target interaction prediction~\cite{gilson2016bindingdb,zitnik2018biosnap,liu2015improving,chen2020transformercpi}, while Davis and KIBA are widely used for binding affinity prediction~\cite{davis2011comprehensive,tang2014making}. Together, these resources have driven progress in structure-based analysis, virtual screening, interaction prediction, and affinity estimation. However, they generally do not provide large-scale residue--atom annotations specifying which protein residues contact the ligand and which non-covalent interactions stabilize these contacts. InteractBind fills this gap by extracting non-covalent interactions from experimentally resolved protein--ligand complexes and converting them into sequence-level residue--atom annotations and aggregated binding-site maps, enabling the systematic training and evaluation of binding-site localization beyond pair-level labels.

\subsection{Protein--ligand evaluation metrics}

Evaluation metrics for protein--ligand modeling typically follow the level of supervision provided by existing datasets. For binary interaction prediction, standard classification metrics such as AUROC, AUPRC, accuracy, precision, recall, and F1 scores are commonly used to assess whether a model distinguishes interacting from non-interacting protein--ligand pairs~\cite{huang2021moltrans,chen2020transformercpi,bai2023interpretable}. For binding affinity prediction, regression and ranking metrics such as mean squared error, root mean squared error, Pearson correlation, Spearman correlation, and concordance index are used to measure agreement between the predicted and experimental affinity values~\cite{davis2011comprehensive,tang2014making, liu2015improving}. Structure-based evaluation commonly relies on geometric criteria, such as ligand RMSD and docking success rate, to assess whether predicted binding poses match reference structures~\cite{trott2010autodock,mysinger2012directory,morehead2025assessing}. However, these metrics primarily evaluate pair-level outcomes or geometric agreement, and do not directly assess whether a model identifies the binding residues on the protein. InteractBind therefore complements conventional outcome-level metrics with binding-site localization metrics based on residue-level hit rates. Specifically, we use Binding Residue Hit Rate at Top-$K$ (BRHR@$K$), which measures whether the top-ranked model-predicted residues overlap with ground-truth binding residues. The full task definition and evaluation protocol are provided in Sec.~\ref{sec:tasks_eval}.

\section{The InteractBind Benchmark}

Fig.~\ref{fig:benchmark_overview} provides an overview of InteractBind, including dataset construction from experimentally resolved protein--ligand complexes and the resulting annotations for each protein--ligand pair. These annotations enable the evaluation of whether model-derived interaction maps localize the protein residues involved in binding. Tab.~\ref{tab:dataset_comparison} further contrasts InteractBind with existing representative protein--ligand datasets, highlighting its support for a broader range of protein--ligand modeling tasks. In the following subsections, we describe the dataset curation procedure (Sec.~\ref{sec:benchmark_design}), summarize the dataset statistics (Sec.~\ref{sec:benchmark_statistics_section}), and define the tasks and evaluation protocol used in this study (Sec.~\ref{sec:tasks_eval}).

\begin{table*}[t]
\centering
\footnotesize
\caption{
Comparison of sequence-based protein--ligand datasets. 
``\checkmark'' indicates that the information is provided or directly supported by the dataset, whereas ``--'' indicates that it is not provided.}
\label{tab:dataset_comparison}

\setlength{\tabcolsep}{6.0pt}
\renewcommand{\arraystretch}{1.12}
\begin{tabular}{lccccc}
\rowcolor{gray!25}
\Xhline{0.8pt}
\textbf{Dataset}
& \textbf{Binary label}
& \textbf{Affinity}
& \textbf{Docking negatives}
& \textbf{Binding sites}
& \textbf{Interaction-type annotations} \\
\Xhline{0.8pt}
DUD-E~\cite{mysinger2012directory} & \checkmark & -- & \checkmark & -- & -- \\
BindingDB~\cite{gilson2016bindingdb} & \checkmark & \checkmark & -- & -- & -- \\
BioSNAP~\cite{zitnik2018biosnap} & \checkmark & -- & -- & -- & -- \\
Human~\cite{liu2015improving,chen2020transformercpi} & \checkmark & -- & -- & -- & -- \\
Davis~\cite{davis2011comprehensive} & -- & \checkmark & -- & -- & -- \\
KIBA~\cite{tang2014making} & -- & \checkmark & -- & -- & -- \\
\textbf{InteractBind} & \checkmark & \checkmark & \checkmark & \checkmark & \checkmark \\
\bottomrule
\end{tabular}
\vspace{-1em}
\end{table*}

\subsection{Dataset curation}
\label{sec:benchmark_design}

InteractBind is curated from experimentally resolved protein--ligand complexes in the Protein Data Bank (PDB), which provide atomic coordinates for both proteins and bound ligands~\citep{berman2007worldwide}. For each protein, we construct both the canonical FASTA sequence and the SaProt structure-aware (SA) sequence~\cite{su2023saprot}; for each ligand, we convert PDB coordinates to SMILES and SELFIES, using SELFIES as a robust ligand sequence representation for sequence-based interaction map modeling~\cite{krenn2022selfies}. Non-covalent residue--atom interactions are detected from each three-dimensional complex using PLIP~\cite{schake2025plip} and GetContacts~\cite{getcontacts}, and projected onto the corresponding protein residue and ligand atom to construct sequence-level interaction maps. Therefore, the resulting sample contains paired protein and ligand sequence representations together with sequence-aligned interaction annotations, as illustrated in Fig.~\ref{fig:benchmark_overview}~A--B. Details of structure collection, filtering, representation conversion, and sample construction are provided in Appendix~\ref{app:structure_collection}. The interaction annotations are constructed from six major non-covalent interaction types: hydrogen bonds, salt bridges, van der Waals interactions, hydrophobic contacts, $\pi$--$\pi$ stacking, and cation--$\pi$ interactions. These interactions represent major physicochemical forces underlying protein--ligand recognition and provide an interpretable basis for describing residue--atom contacts~\cite{vcerny2007non,holehouse2024molecular}. Interaction validity is determined using rule-based geometric and chemical constraints, including distance cutoffs and, where applicable, angular conditions. The six interaction-specific maps are also aggregated into an overall interaction map, from which protein binding-site annotations are derived: a residue is labeled as part of the binding site if it participates in at least one detected ligand-stabilizing non-covalent interaction. The full interaction annotation protocol and interaction-specific cutoffs are provided in Appendix~\ref{app:interaction_annotation} and Tab.~\ref{plip-key-cutoffs}.

\subsection{Dataset statistics}
\label{sec:benchmark_statistics_section}

\begin{figure*}[htbp]
    \centering
    \includegraphics[width=1\textwidth]{ 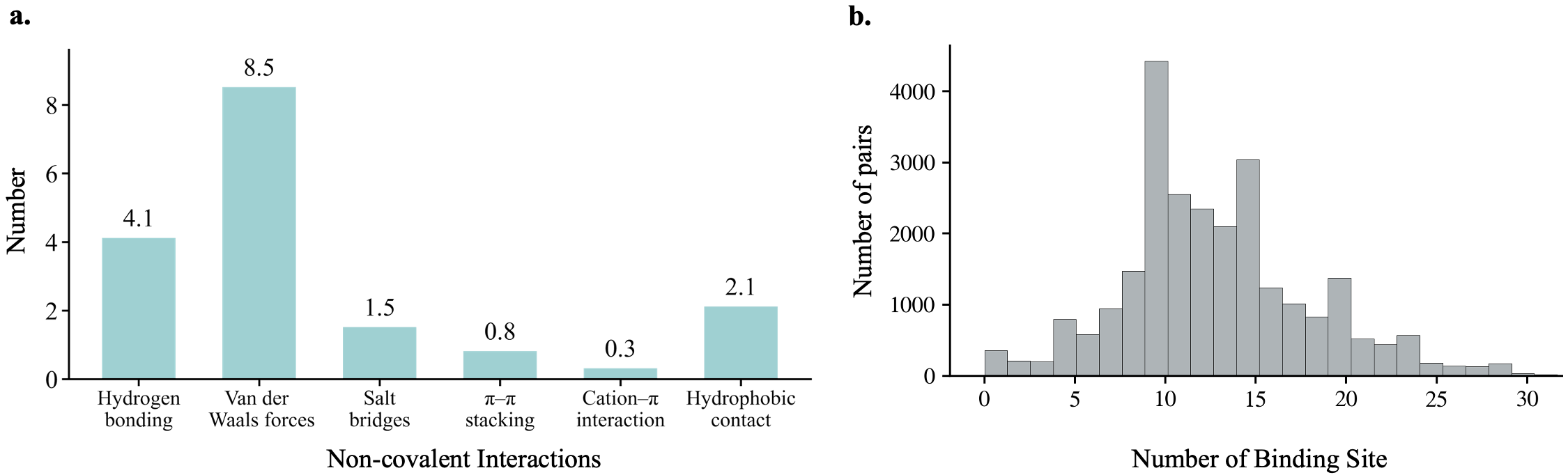}
    \caption{
    \textbf{Descriptive statistics of InteractBind.}
    \textbf{a.} Average numbers of the six annotated non-covalent interaction types across InteractBind.
    \textbf{b.} Distribution of annotated binding-site residues across InteractBind.
    }
    \label{fig:benchmark_statistics}
    \vspace{-1em}
\end{figure*}

InteractBind contains 99,391 protein--ligand pairs, covering 11,473 unique proteins and 9,017 unique ligands. Each sample includes paired protein and ligand sequence representations, binary binding labels, docking-based affinity values, six non-covalent interaction maps, and an aggregated binding-site map, as summarized in Fig.~\ref{fig:benchmark_overview}~B. To support both ID and OOD evaluation, we further construct an affinity-based ID subset and four protein similarity-controlled OOD splits. The detailed statistics for these datasets are provided in Tab.~\ref{tab:interactbind_split_statistics}. Fig.~\ref{fig:benchmark_statistics} summarizes the interaction-level and binding-site statistics of InteractBind. Panel~a shows that the dataset is dominated by common protein--ligand contact patterns, including van der Waals interactions, hydrogen bonds and hydrophobic contacts, while also retaining less frequent but chemically informative interaction types such as salt bridges, $\pi$--$\pi$ stacking and cation--$\pi$ interactions. This distribution reflects the natural imbalance of non-covalent contacts in protein--ligand complexes and motivates interaction-type-specific evaluation rather than treating all contacts as a single homogeneous signal. Panel~b further shows that annotated binding sites vary substantially in size, with most complexes containing a moderate number of binding-site residues and a long tail of larger sites. This diversity indicates that InteractBind covers binding interfaces of different extents, providing a basis for evaluating whether models can localize binding residues beyond simple or narrowly defined pocket patterns. Additional statistics, including protein sequence lengths, ligand lengths, and molecular weight distributions, are provided in Fig.~\ref{fig:main_fig_distribution}.

\section{Tasks and Evaluation Protocol}
\label{sec:tasks_eval}

\noindent\textbf{Dataset splits.}
InteractBind supports both in-distribution and out-of-distribution evaluation. For the in-distribution setting, we adopt an affinity-aware protocol because defining positive and negative pairs using docking-based affinity scores is more reliable than constructing negatives solely by random pairing. Specifically, pairs with affinity scores $\leq -7$ are treated as high-confidence positive samples, pairs with affinity scores $> -5$ are treated as high-confidence negative samples, and intermediate cases are excluded to maintain a clear margin between the two classes and reduce ambiguity from docking-score noise. The resulting subset is evaluated using 5-fold cross-validation, and results are reported as mean $\pm$ standard deviation across folds. 

For out-of-distribution evaluation, we construct protein similarity-controlled splits because this setting more closely reflects the practical scenario of predicting interactions for less familiar protein targets~\cite{jia2026deep}. In sequence-based protein--ligand modeling, the difficulty of generalization is strongly affected by the similarity between training and test sequences, whereas cold-start partitioning based only on identifiers or names does not necessarily define a more challenging or realistic setting~\cite{chatterjee2023improving,jia2026deep}. Protein similarity is measured using global sequence alignment based on the Needleman--Wunsch algorithm~\cite{hamamsy2024protein}. This approach yields four protein-based OOD datasets with maximum train--test protein similarities of 25\%, 28\%, 31\%, and 33\%. Each dataset is divided into training, validation, and test subsets with a ratio of 8:1:1. The reported OOD results are obtained over 5 runs with different random seeds and are presented as mean $\pm$ standard deviation. These splits provide a controlled setting for evaluating binary binding prediction and binding-site localization under target-side distribution shift.

\noindent\textbf{Baseline models.}
We evaluate eight existing representative sequence-based protein--ligand models whose interaction modules have been qualitatively used to support claims of binding-site localization. Because these models produce learned residue--atom interaction maps in addition to binary binding predictions, InteractBind enables us to test these claims systematically by comparing model-derived maps against ground-truth binding-site annotations. The evaluated models include transformer- or cross-attention-based architectures, including MolTrans~\cite{huang2021moltrans}, TransformerCPI~\cite{chen2020transformercpi}, HyperAttentionDTI~\cite{zhao2022hyperattentiondti}, PerceiverCPI~\cite{nguyen2023perceiver}, CAT-DTI~\cite{zeng2024cat}, and FusionDTI~\cite{meng-etal-2025-fusiondti}, and bilinear-attention-network-based architectures, including DrugBAN~\cite{bai2023interpretable} and GraphBAN~\cite{hadipour2025graphban}. Detailed descriptions of the evaluated baselines are provided in Appendix~\ref{baseline_detail}.

\noindent\textbf{Binary binding prediction evaluation.}
For binary binding prediction, we follow standard metrics used in protein--ligand interaction modeling, including AUROC, AUPRC, accuracy, and F1 score~\cite{hadipour2025graphban,meng-etal-2025-fusiondti}. These metrics quantify whether a model predicts binding correctly, but they do not indicate whether the model localizes the protein residues involved in ligand recognition.

\noindent\textbf{Binding-site localization evaluation.}
For binding-site localization, we define a Top-$K$ metric termed the \textit{Binding Residue Hit Rate} (BRHR). For a given protein--ligand pair, let $\mathbf{P} \in \mathbb{R}^{m \times n}$ denote the model-derived interaction map, where $m$ is the number of ligand tokens or atoms, $n$ is the number of protein residues, and each entry $P_{ij}$ indicates the predicted interaction score between ligand unit $i$ and protein residue $j$. Because BRHR focuses on protein-side binding-site localization, we aggregate the interaction map along the ligand dimension to obtain a residue-level score:
\begin{equation}
    s_j = \max_{i \in \{1,\ldots,m\}} P_{ij},
\end{equation}
where $s_j$ denotes the predicted binding-site score for protein residue $j$. We then rank all protein residues by $s_j$ and define:
\begin{equation}
    R_K = \operatorname{Top-}K(\{s_j\}_{j=1}^{n})
\end{equation}
as the set of Top-$K$ predicted protein residues. Let $T$ denote the set of ground-truth binding-site residues. We define BRHR for a single sample as follows:
\begin{equation}
\mathrm{BRHR}(K)=
\begin{cases}
1, & \text{if } R_K \cap T \neq \varnothing,\\
0, & \text{otherwise.}
\end{cases}
\end{equation}
In other words, a prediction is counted as correct if at least one of the Top-$K$ predicted protein residues matches a ground-truth binding-site residue. The dataset-level score is computed as:
\begin{equation}
\mathrm{BRHR}_{\mathrm{avg}}(K)=\frac{1}{N}\sum_{i=1}^{N}\mathrm{BRHR}_i(K), \quad \mathrm{BRHR}_{\mathrm{avg}}(K)\in[0,1],
\end{equation}
where $N$ is the number of evaluated protein--ligand pairs. Larger values indicate better binding-site localization performance.

\noindent\textbf{Evaluation scope.}
We use BRHR to evaluate binding-site localization across models with different interaction granularities. Although InteractBind provides ligand-atom $\times$ protein-residue interaction maps, sequence-based models use ligand atoms, substructures, graph nodes, or learned tokens as their basic ligand units, making direct atom--residue contact prediction difficult to compare fairly across models. Thus, we project model-derived interaction maps onto the protein-residue axis and assess whether they recover annotated binding residues.

\section{Results}

\subsection{From binary binding prediction to binding-site localization}

A central goal of InteractBind is to evaluate whether protein--ligand models can move beyond binary binding prediction and localize the binding sites that support ligand recognition. We therefore examine eight representative interaction-aware baselines on both binary binding prediction and binding-site localization under the in-distribution split. As shown in Tab.~\ref{tab:interactbind_id}, models with stronger binary binding prediction generally also achieve stronger binding-site localization, suggesting a positive association between the two tasks on the InteractBind benchmark. FusionDTI performs best across both tasks, achieving 98.3\% AUROC and 90.1\% accuracy for binary binding prediction, together with 21.6\% BRHR@1, 29.6\% BRHR@3, and 35.6\% BRHR@5 for binding-site localization. This is consistent with FusionDTI's token-level fusion design, which explicitly models residue--atom correspondences. Bilinear-attention models also show competitive localization performance, with GraphBAN reaching 16.1\% BRHR@1 and DrugBAN reaching 31.4\% BRHR@5. These results indicate that learned protein--ligand interaction maps contain measurable binding-site signals, especially when models explicitly learn residue--atom interactions. Nevertheless, binding-site localization remains substantially underdeveloped across the existing baselines. Indeed, even FusionDTI reaches only 21.6\% BRHR@1 despite achieving 98.3\% AUROC, showing that strong binary binding prediction does not yet translate into reliable top-ranked binding-site localization. This limitation is expected because these baselines are trained primarily with binary binding labels, without large-scale binding-site annotations to guide their interaction maps. InteractBind therefore provides the supervision and evaluation setting needed to assess whether models localize binding sites, rather than only whether they predict binding correctly.

\begin{table*}[t]
\centering
\footnotesize
\caption{Performance (\%) of binary binding prediction and binding-site localization under the ID setting. The best results are shown in \textbf{bold}, and the second-best results are \underline{underlined}. \textbf{BRHR} denotes the binding residue hit rate, which measures whether one of the Top-$K$ residues matches a ground-truth binding residue, with larger values indicating better binding-site localization.}
\label{tab:interactbind_id}

\setlength{\tabcolsep}{5.6pt}
\renewcommand{\arraystretch}{1.05}

\begin{tabular}{lcccc|ccc}
\rowcolor{gray!25}
\Xhline{0.8pt}
\textbf{Model} 
& \multicolumn{4}{c|}{\textbf{Binary binding prediction}} 
& \multicolumn{3}{c}{\textbf{Binding-site localization}} \\
\Xhline{0.8pt}
\rowcolor{gray!15}
& \textbf{AUROC} & \textbf{AUPRC} & \textbf{ACC} & \textbf{F1} 
& \textbf{BRHR@1} & \textbf{BRHR@3} & \textbf{BRHR@5} \\
\Xhline{0.8pt}
MolTrans          
& 95.1 $\pm$ 2.2
& 93.7 $\pm$ 1.2
& 88.6 $\pm$ 0.4
& 88.1 $\pm$ 0.5
& 10.9 $\pm$ 0.5
& 17.7 $\pm$ 0.2
& 22.4 $\pm$ 0.8 \\

TransformerCPI    
& 95.4 $\pm$ 0.3
& 90.8 $\pm$ 0.8
& 89.1 $\pm$ 0.4
& 88.5 $\pm$ 0.6
& 10.9 $\pm$ 0.2
& 16.7 $\pm$ 0.7
& 21.7 $\pm$ 0.4 \\

HyperAttentionDTI 
& 95.5 $\pm$ 0.6
& 92.8 $\pm$ 0.3
& 84.1 $\pm$ 0.9
& 88.5 $\pm$ 0.5
& 12.3 $\pm$ 0.4
& 17.7 $\pm$ 0.6
& 20.8 $\pm$ 0.3 \\

PerceiverCPI      
& 96.9 $\pm$ 0.2
& 92.6 $\pm$ 0.5
& 86.6 $\pm$ 0.8
& 82.2 $\pm$ 0.7
& 10.6 $\pm$ 0.1
& 13.9 $\pm$ 0.5
& 19.9 $\pm$ 0.9 \\

CAT-DTI           
& 96.8 $\pm$ 0.4
& 92.5 $\pm$ 0.6
& 88.1 $\pm$ 0.5
& 87.0 $\pm$ 0.3
& 12.2 $\pm$ 0.7
& 14.8 $\pm$ 0.2
& 22.1 $\pm$ 0.6 \\

DrugBAN           
& 97.6 $\pm$ 1.2
& 94.3 $\pm$ 0.4
& \underline{89.3 $\pm$ 0.4}
& 88.3 $\pm$ 1.4
& 15.7 $\pm$ 0.6
& \underline{23.9 $\pm$ 0.1}
& \underline{31.4 $\pm$ 0.4} \\

GraphBAN          
& \underline{97.8 $\pm$ 0.2}
& \underline{94.4 $\pm$ 0.3}
& 88.9 $\pm$ 0.4
& \underline{88.5 $\pm$ 0.4}
& \underline{16.1 $\pm$ 0.5}
& 22.9 $\pm$ 0.2
& 30.4 $\pm$ 0.6 \\

FusionDTI         
& \textbf{98.3 $\pm$ 0.2}
& \textbf{95.9 $\pm$ 0.4}
& \textbf{90.1 $\pm$ 0.3}
& \textbf{89.9 $\pm$ 0.2}
& \textbf{21.6 $\pm$ 0.1}
& \textbf{29.6 $\pm$ 0.3}
& \textbf{35.6 $\pm$ 0.2} \\

\bottomrule
\end{tabular}
\vspace{-1em}
\end{table*}

\subsection{Per-interaction-type analysis of binding-site localization}

\begin{figure*}[t]
    \centering
    \includegraphics[width=0.9\textwidth]{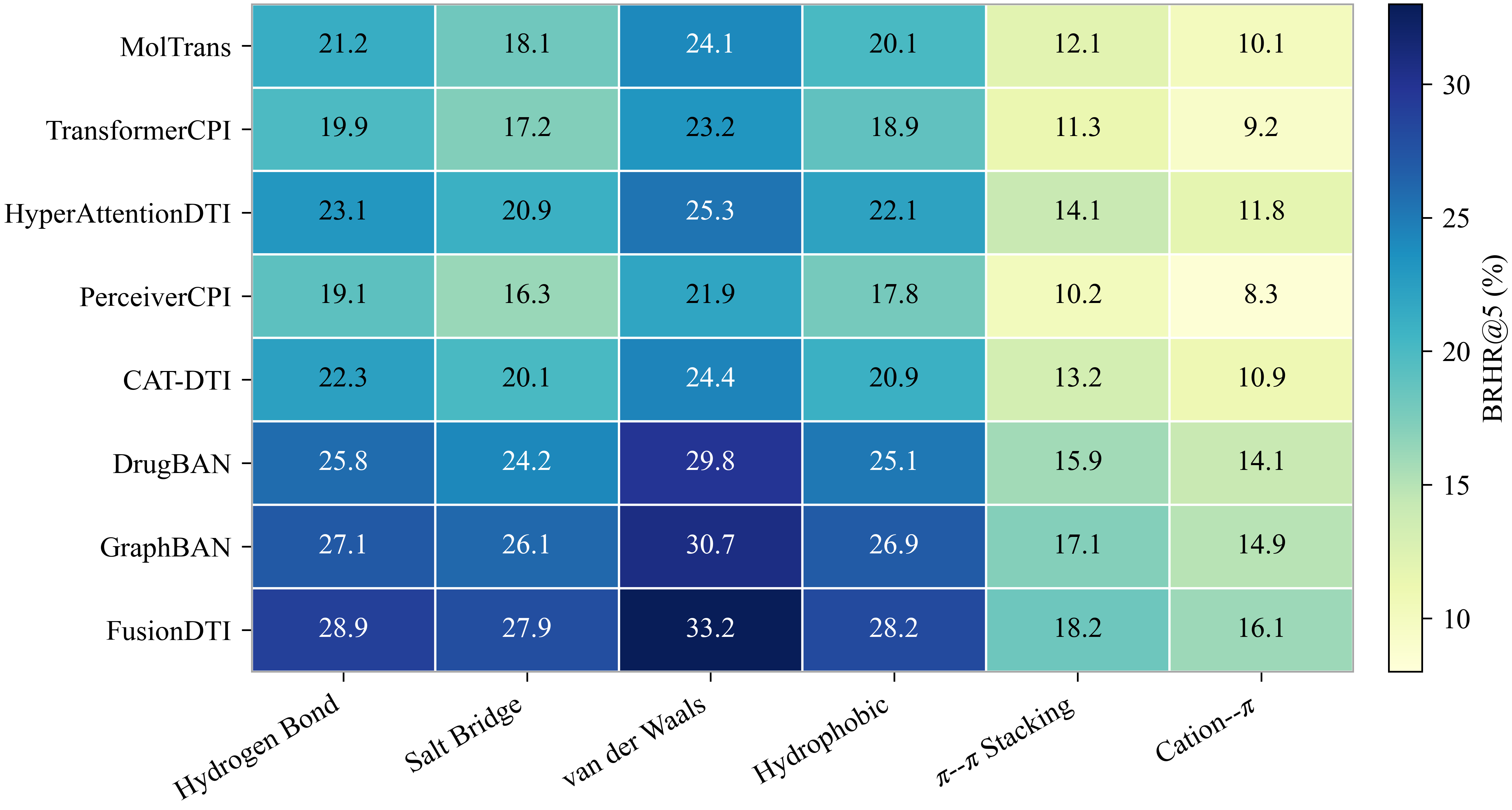}
    \caption{
    \textbf{Interaction-type-specific binding-site localization in the in-distribution setting.}
    Heatmap of BRHR@5 (\%) across eight representative models and six non-covalent interaction types.
    }
    \label{fig:interaction_type_brhr5}
    \vspace{-1em}
\end{figure*}

A distinctive feature of InteractBind is that binding-site annotations are decomposed into six non-covalent interaction types, enabling a fine-grained evaluation beyond an aggregate binding-site map. We therefore evaluate binding-site localization separately for each interaction category. As shown in Fig.~\ref{fig:interaction_type_brhr5}, BRHR@5 varies substantially across interaction types under exact-hit matching. Van der Waals interactions are the easiest to localize, with FusionDTI reaching 33.2\% BRHR@5, followed by GraphBAN at 30.7\% and DrugBAN at 29.8\%. This trend is consistent with the relatively common and spatially broad nature of van der Waals contacts, which provide more diffuse residue-level localization signals than highly specific interaction modes. Hydrophobic contacts and hydrogen bonds also show relatively strong localization performance, with FusionDTI achieving 28.2\% and 28.9\% BRHR@5, respectively. Salt bridges follow a similar pattern, with FusionDTI reaching 27.9\% and GraphBAN 26.1\%. By contrast, $\pi$--$\pi$ stacking and cation--$\pi$ interactions remain more challenging: the best BRHR@5 values are 18.2\% for $\pi$--$\pi$ stacking and 16.1\% for cation--$\pi$, both achieved by FusionDTI. This is expected because aromatic and electrostatic interactions require more specific geometric arrangements and residue--atom alignment, making exact-hit localization more difficult. These results indicate that binding-site localization is not uniform across physicochemical regimes. Common and spatially broader contacts, such as van der Waals, hydrophobic, and hydrogen-bond interactions, are more readily localized, whereas aromatic, electrostatic, and geometry-specific interactions remain harder to recover. Although models with stronger overall binding-site localization also tend to perform better across interaction types, the low scores for $\pi$--$\pi$ stacking and cation--$\pi$ show that fine-grained non-covalent interaction patterns remain difficult to identify reliably. Thus, the interaction-type-specific annotations in InteractBind provide a diagnostic benchmark signal that aggregate binding-site localization alone cannot capture, revealing which physicochemical contact patterns are localized well and which remain poorly resolved.

\subsection{Generalization towards novel proteins}

\begin{figure*}[t]
    \centering
    \includegraphics[width=0.95\textwidth]{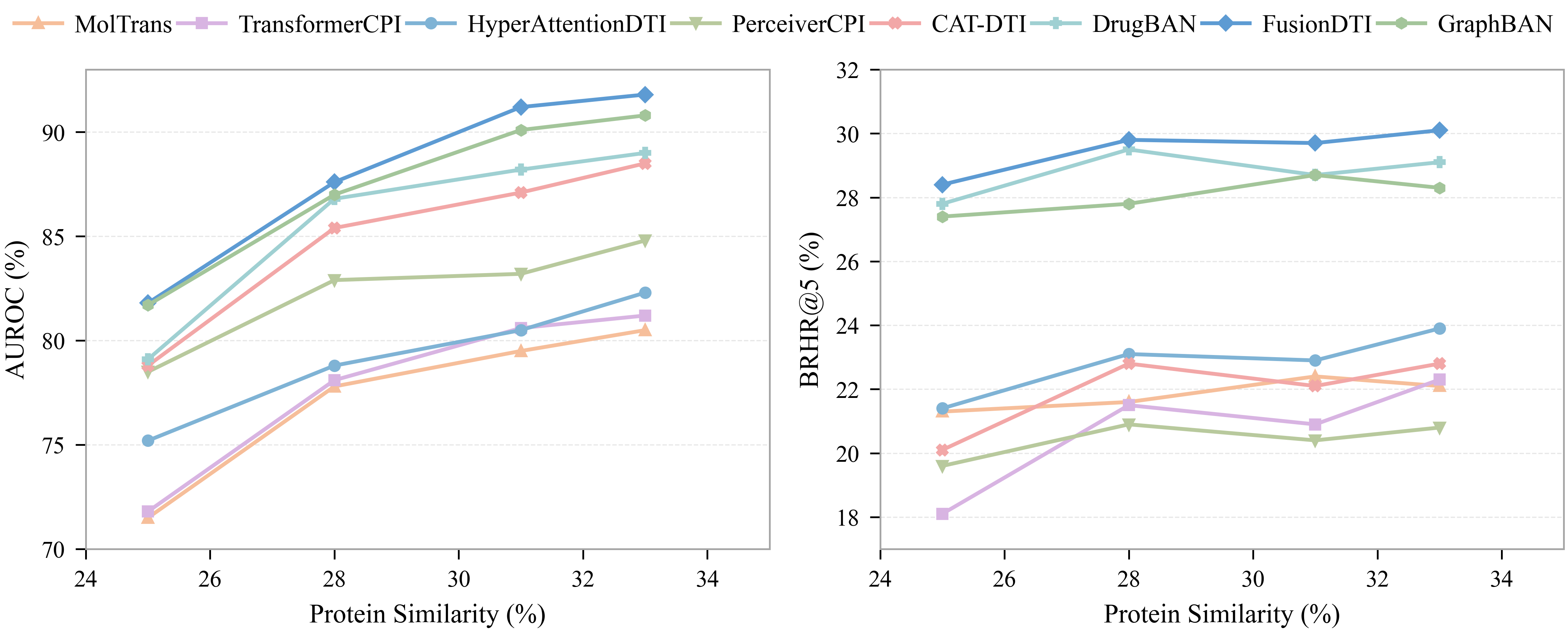}
    \caption{
    \textbf{Generalisation across protein similarity-controlled OOD datasets.}
    The four splits constrain maximum train--test protein sequence similarity to 25\%, 28\%, 31\%, and 33\%.
    \textbf{Left:} binary binding prediction performance measured by AUROC.
    \textbf{Right:} binding-site localisation performance measured by BRHR@5.
    }
    \label{fig:protein_similarity_results}
    \vspace{-1em}
\end{figure*}

A key goal of InteractBind is to support realistic generalization assessment beyond random or in-distribution evaluation. To this end, we construct four protein similarity-controlled datasets, enabling controlled evaluation of model inference on novel protein targets. As shown in Fig.~\ref{fig:protein_similarity_results}~(left), binary binding prediction improves as the protein similarity constraint becomes less stringent. FusionDTI increases from 81.8\% AUROC at 25\% similarity to 91.8\% at 33\%, while GraphBAN improves from 81.7\% to 90.8\% and DrugBAN from 79.1\% to 89.0\%. A similar trend is observed for other baselines. This trend is expected for sequence-based models, as test proteins that are more similar to training proteins provide more familiar sequence patterns and reduce target-side distribution shift. InteractBind makes this effect explicit by quantifying performance across controlled protein similarity thresholds. Binding-site localization shows a similar but more moderate trend. As shown in Fig.~\ref{fig:protein_similarity_results}~(right), FusionDTI 's performance increases from 28.4\% BRHR@5 at 25\% similarity to 30.1\% at 33\%, while it ranges from 27.8\% to 29.1\% for DrugBAN. In contrast, other baselines generally remain below 24\% BRHR@5 across the four datasets. These findings indicate that train--test protein similarity affects not only binary binding prediction but also binding-site localization. Thus, the protein similarity-controlled splits in InteractBind provide a more informative stress test for evaluating whether models can localize binding sites on less familiar protein targets. We also construct ligand similarity-controlled splits and observe a smaller effect on model performance, with detailed results provided in Appendix~\ref{app:ligand_similarity_results}.

\section{Discussion}

The results on InteractBind support three main conclusions. \textbf{First}, current interaction-aware models show measurable binding-site localization ability on the InteractBind benchmark, suggesting that their learned protein--ligand interaction maps capture useful binding-site signals. Moreover, models with stronger binary binding prediction generally achieve better binding-site localization, indicating a positive association between the two tasks. However, even the best-performing model, FusionDTI, reaches only 21.6\% BRHR@1 despite achieving 98.3\% AUROC, showing that substantial room remains for improving binding-site localization. This limitation likely reflects the current reliance on binary binding labels, since it is difficult without large-scale binding-site supervision to directly guide model-derived interaction maps. \textbf{Second}, per-interaction-type evaluation shows that binding-site localization is not a uniform capability of the evaluated models. Although the overall binding-site map aggregates all non-covalent contacts, the residues in this map arise from interaction types with different physicochemical constraints. Common and spatially diffuse contacts, such as hydrophobic and van der Waals interactions, are generally easier to localize, whereas contacts requiring more specific aromatic, electrostatic, or geometric arrangements remain more challenging. This suggests that aggregate BRHR scores alone can mask important differences in how models handle distinct residue--atom contact patterns. Interaction-type-specific annotations therefore provide a more diagnostic evaluation of whether models identify the physicochemical interactions that stabilize ligand binding. \textbf{Third}, protein similarity-controlled OOD evaluation shows that train--test protein similarity is an important factor affecting model inference on unseen protein--ligand combinations. As the maximum train--test protein similarity becomes less constrained, binary binding prediction improves, and binding-site localization shows a similar trend. These results indicate that protein similarity between training and test data strongly influences model performance, highlighting the need for similarity-controlled splits in realistic generalization assessment. Overall, InteractBind complements conventional protein--ligand benchmarks by adding large-scale binding-site and interaction-type-specific annotations derived from non-covalent contacts in experimentally resolved complexes. It enables a systematic evaluation of whether the models localize binding sites and identify the interaction patterns that stabilize ligand binding, providing a foundation for future models with stronger fine-grained localization ability.

\noindent\textbf{Broader impact.}
Protein--ligand binding prediction underpins computational drug discovery and molecular design, yet strong binary binding prediction does not necessarily indicate that models can localize binding sites or identify the interaction patterns governing molecular recognition. Models may instead exploit dataset-specific correlations, leading to unreliable predictions under distribution shift. By supporting a systematic evaluation of binding-site localization and interaction-type-specific contact identification, InteractBind encourages a shift from outcome-only evaluation towards fine-grained assessment of protein--ligand models. This may promote the development of more physically interpretable methods and improve the reliability of computational drug discovery pipelines, where incorrect binding-site or interaction-pattern inference can lead to misleading biological conclusions or suboptimal compound prioritization.

\noindent\textbf{Limitations.}
The current version of InteractBind focuses on sequence-level evaluation and does not yet include a structure-level benchmark. Extending the evaluation to three-dimensional structures is non-trivial, since protein--ligand binding often admits multiple plausible poses and involves conformational flexibility that is not captured by a single static structure. This ambiguity makes it difficult to define a consistent and reliable ground truth for structure-level comparison. As a result, structure-based evaluation may introduce additional noise and reduce interpretability compared to residue-level sequence-based assessment. Developing a robust and standardized protocol for structure-level benchmarking remains an important and challenging direction for future work.

\clearpage

{\small
\bibliographystyle{unsrtnat}
\bibliography{refs}
}

\clearpage
\appendix

\section{Preliminaries}
\label{sec:preliminary}

\paragraph{Covalent bonds and non-covalent interactions.}
Molecular systems involve two fundamental types of interactions: \textit{covalent bonds} and \textit{non-covalent interactions}~\cite{vcerny2007non}. Covalent bonds govern intra-molecular organization and are typically treated as fixed in modeling. In contrast, non-covalent, generally reversible interactions, including hydrogen bonds~\cite{murai1993efficient}, salt bridges~\cite{pylaeva2018salt}, van der Waals contacts~\cite{yang2013much}, hydrophobic contacts~\cite{sitarik2025widespread}, $\pi$--$\pi$ stacking~\cite{carter2020reinterpreting}, and cation--$\pi$ interactions~\cite{dougherty2013cation}, do not involve bond formation but arise from physicochemical compatibility between molecules. These interactions are predominantly inter-molecular and collectively determine protein--ligand binding, including affinity, specificity, and binding-site localization. However, such local mechanistic details are typically absent from existing datasets. InteractBind addresses this limitation by explicitly representing such local interaction patterns through sequence-level interaction maps.

\paragraph{Binding-site and binding pocket.}
In structural biology, a \textit{binding pocket} generally refers to a broader three-dimensional region on the protein surface that can accommodate a ligand, often characterized by geometric cavities or surface concavities~\cite{liang1998anatomy}. In contrast, a \textit{binding-site} is more specific and denotes the subset of protein residues that directly participate in ligand recognition through non-covalent interactions~\cite{mattos1996locating}. While binding pockets describe a spatial region that may potentially host a ligand, binding sites capture the actual interacting components that determine molecular recognition and binding specificity. Although the two concepts are closely related and sometimes used interchangeably, they differ in granularity and interpretability. Binding pocket identification is often based on structural geometry and provides a coarse description of possible ligand-accessible regions, whereas binding-site annotation reflects the underlying specific physicochemical interactions that stabilize binding~\cite{stank2016protein}. In many sequence-based settings where high-resolution structural information is limited, binding pocket definitions can be ambiguous, making it difficult to evaluate whether a model has learned the true interaction determinants~\cite{masters2025investigating}. In this work, we therefore focus on binding sites at the residue level, defined by experimentally-derived interaction annotations. This formulation provides a more precise and mechanistically grounded target for evaluation, enabling a fine-grained assessment of whether models localize the residues involved in protein--ligand recognition.



\section{Compute Resources}

Dataset construction and all baseline experiments were conducted on our institutional computing cluster equipped with NVIDIA GH200 GPUs.

\section{Curation of the InteractBind Dataset}
\label{app:construction_details}

\subsection{Structure Collection and Sample Construction.}
\label{app:structure_collection}

We construct \textbf{InteractBind}, a large-scale non-covalent interaction supervised protein--ligand binding dataset, by collecting protein--ligand complex structures from the Protein Data Bank (PDB) format (version 3.2) using PyMOL~\cite{py3dmol}. Structures were filtered to retain only those that contain at least one organic ligand that binds to the protein. Each retained structure underwent a standardized pre-processing pipeline. All inorganic ligands and crystallographic water molecules were removed. Moreover, complexes containing metal ions in either the ligand or the protein were excluded. If a PDB entry contained multiple organic ligands, distance-based filtering was applied to extract individual ligand--protein complexes: for each ligand, only protein chains within $0.5\,\mathrm{nm}$ of the ligand were retained, while others were discarded. As a result, PDB files containing multiple ligands were split into multiple complexes, each consisting of a single ligand and its interacting protein chain(s). 

For each complex, the input data were then separated into two files while preserving the original atomic coordinates. Protein structures were converted to structure-aware (SA)/FASTA format. We derive structure-aware (SA) protein representations via Foldseek~\cite{van2024fast}, providing an explicit local structural context~\cite{su2023saprot}. Ligands were converted from PDB to SMILES and subsequently to SELFIES representations. To obtain labeled binding data, a docking-based strategy was employed. Focused docking was performed for each ligand--protein pair, with the search box centered on the original ligand position and each axis set to the ligand size plus a $0.5\,\mathrm{nm}$ buffer. Pairs with predicted binding affinities lower (i.e., stronger) than $-7.0\,\mathrm{kcal/mol}$ were labeled as positive samples. Negative samples were generated by randomly selecting ligands for each positive protein and performing global docking with the search box centered at the protein centroid and sized as the protein dimensions plus a $1.0\,\mathrm{nm}$ buffer. Pairs with top-ranked docking affinities higher (i.e., weaker) than $-5.0\,\mathrm{kcal/mol}$ were labeled as negative samples. This procedure was repeated until each positive protein was associated with at least one valid negative ligand.

\begin{figure}[htbp]
    \centering
    \includegraphics[width=0.8\linewidth]{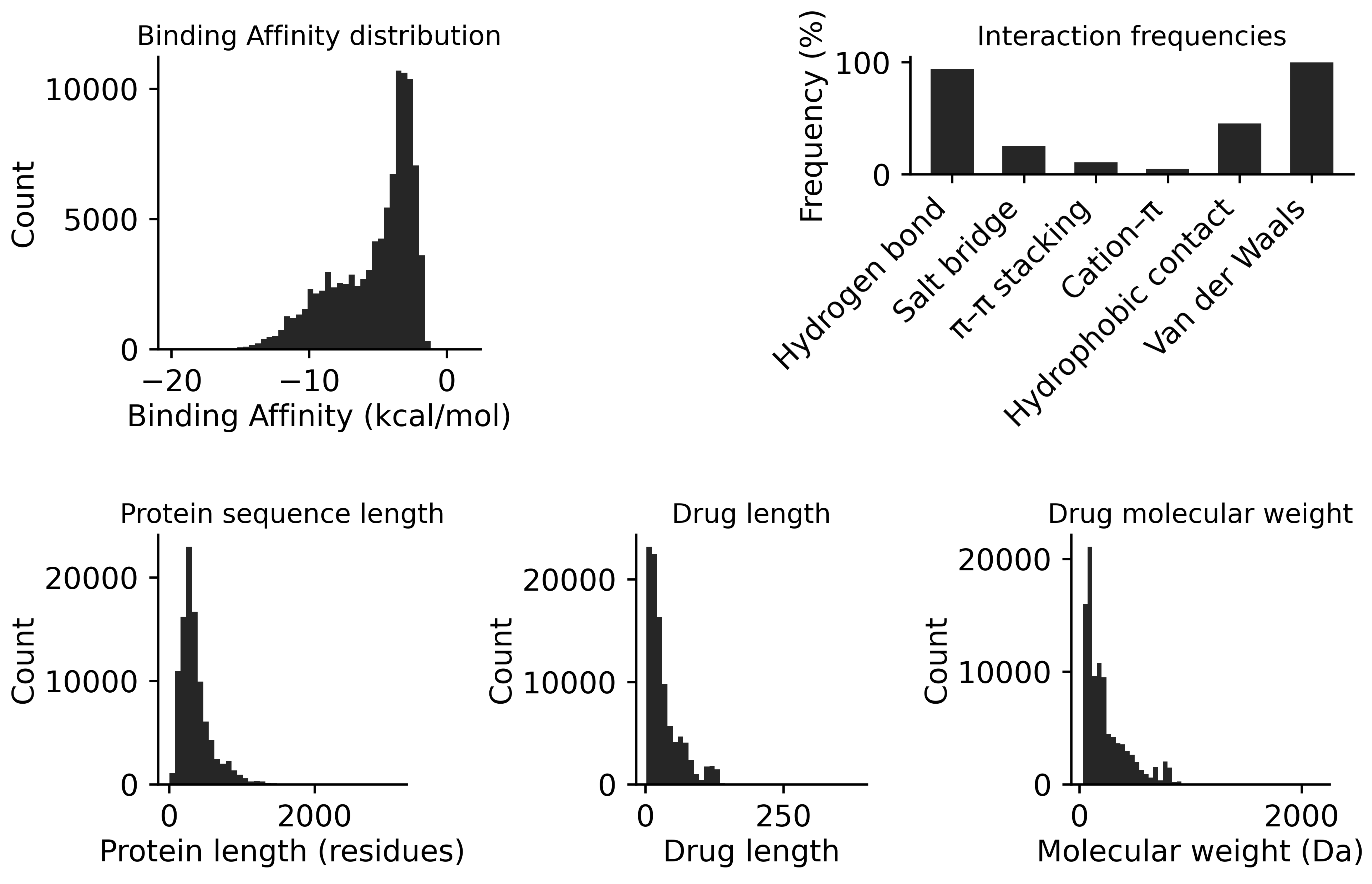}
    \caption{Distribution of features in \textbf{InteractBind}, including binding affinities, non-covalent interactions, protein sequence lengths, ligand lengths, and molecular weights.}
    \label{fig:main_fig_distribution}
\end{figure}

\begin{table*}[t]
\centering
\footnotesize
\caption{
Statistics of the complete InteractBind dataset, the in-distribution subset, and the protein similarity-controlled OOD datasets.
}
\label{tab:interactbind_split_statistics}

\setlength{\tabcolsep}{6.0pt}
\renewcommand{\arraystretch}{1.12}

\begin{tabular}{lllrcc}
\toprule
\textbf{Dataset}
& \textbf{Setting}
& \textbf{Split criterion}
& \textbf{Pairs}
& \textbf{Unique proteins}
& \textbf{Unique ligands} \\
\midrule

InteractBind
& Complete dataset
& All curated pairs
& 99,391
& 11,473
& 9,017 \\

InteractBind-ID
& In-distribution
& Affinity-based split
& 84,825	
& 10,942
& 6,911 \\

InteractBind-P-25\% OOD
& Out-of-distribution
& Protein similarity
& 13,126
& 3,278
& 9,527 \\

InteractBind-P-28\% OOD
& Out-of-distribution
& Protein similarity
& 13,126
& 2,966
& 8,957 \\

InteractBind-P-31\% OOD
& Out-of-distribution
& Protein similarity
& 13,126
& 3,202
& 10,083 \\

InteractBind-P-33\% OOD
& Out-of-distribution
& Protein similarity
& 13,126
& 4,807
& 11,084 \\

\bottomrule
\end{tabular}
\end{table*}

\subsection{Interaction Annotation and Strength Modeling}
\label{app:interaction_annotation}

Protein--ligand binding is governed by a diverse set of non-covalent interactions, including hydrogen bonds~\cite{murai1993efficient}, salt bridges~\cite{pylaeva2018salt}, van der Waals contacts~\cite{yang2013much}, hydrophobic contacts~\cite{sitarik2025widespread}, $\pi$--$\pi$ stacking~\cite{carter2020reinterpreting}, and cation--$\pi$ interactions~\cite{dougherty2013cation}. 
These interactions jointly determine binding affinity, specificity, and structural complementarity, forming the physicochemical basis of molecular recognition~\cite{holehouse2024molecular}. Protein--ligand interactions in \textbf{InteractBind} were detected using rule-based criteria grounded in geometric and chemical constraints. Hydrogen bonds, salt bridges, $\pi$--$\pi$ stacking, cation--$\pi$ interactions, and hydrophobic contacts were identified using PLIP~\cite{schake2025plip}. Van der Waals interactions, which are not captured by PLIP, were computed using GetContacts~\cite{getcontacts}, and only attractive Van der Waals interactions were retained. To quantify interaction strength, we adopt a distance-dependent piecewise linear decay function that maps interatomic or geometric distances to continuous values in the range $[10^{-6}, 1]$ within interaction-specific cutoff windows. Distance thresholds and geometric constraints for each non-covalent interaction are summarized in Tab.~\ref{plip-key-cutoffs}. Angular and geometric criteria, when applicable, are used as binary filters, while the distance-based decay controls the interaction strength.

Based on the detected interactions and their distance-dependent strengths, interaction-supervised attention maps were constructed at the sequence-level. For each protein--ligand complex in \textbf{InteractBind}, an $m \times n$ matrix was generated, where $m$ and $n$ denote the numbers of ligand atoms and protein FASTA residues, respectively. Each matrix entry encodes the interaction strength between a ligand atom and a protein residue. For each complex, six attention maps corresponding to individual non-covalent interactions were generated, together with an additional unified map obtained by summing all interaction-specific maps. All attention maps were stored using unique identifiers for efficient retrieval during training. After cleaning and pre-processing, \textbf{InteractBind} contains a total of \textbf{99,391 protein--ligand pairs}. These complexes form the basis for subsequent analyses and reflect a substantial diversity in binding affinities, interaction patterns, and molecular properties, as shown in Fig.~\ref{fig:main_fig_distribution}.

\begin{table*}[htbp]
\centering
\caption{Geometric criteria and strength definitions for key non-covalent interactions in \textbf{InteractBind}.}
\vspace{-0.4em}
\label{plip-key-cutoffs}
\setlength{\tabcolsep}{11pt}
\renewcommand{\arraystretch}{1.5}
\resizebox{\linewidth}{!}{
\begin{tabular}{llll}
\Xhline{0.8pt}
\rowcolor{gray!20}
\textbf{Non-covalent interaction} & \textbf{Parameter} & \textbf{Geometric criterion} & \textbf{Strength $s(d)$} \\
\Xhline{0.8pt}

Hydrogen bond
& Donor--acceptor distance
& $2.2\,\text{\AA}<d\leq 4.1\,\text{\AA}$
& $
s(d)=
\begin{cases}
\max\!\left(10^{-6},\,1-\dfrac{d-2.2\,\text{\AA}}{4.1\,\text{\AA}-2.2\,\text{\AA}}\right), & 2.2\,\text{\AA}<d\leq 4.1\,\text{\AA} \\
10^{-6}, & \text{otherwise}
\end{cases}
$ \\
& Donor angle & $\geq 100^{\circ}$ & (angle used as filter; distance controls $s$) \\
\hline

Salt bridge
& Distance between charge centres
& $2.8\,\text{\AA}<d\leq 5.5\,\text{\AA}$
& $
s(d)=
\begin{cases}
\max\!\left(10^{-6},\,1-\dfrac{d-2.8\,\text{\AA}}{5.5\,\text{\AA}-2.8\,\text{\AA}}\right), & 2.8\,\text{\AA}<d\leq 5.5\,\text{\AA} \\
10^{-6}, & \text{otherwise}
\end{cases}
$ \\
\hline

$\pi$--$\pi$ stacking
& Ring centroid distance
& $3.4\,\text{\AA}<d\leq 5.5\,\text{\AA}$
& $
s(d)=
\begin{cases}
\max\!\left(10^{-6},\,1-\dfrac{d-3.4\,\text{\AA}}{5.5\,\text{\AA}-3.4\,\text{\AA}}\right), & 3.4\,\text{\AA}<d\leq 5.5\,\text{\AA} \\
10^{-6}, & \text{otherwise}
\end{cases}
$ \\
& Angle deviation & $\leq 30^{\circ}$ & (geometry filter) \\
& Ring offset & $\leq 2.0\,\text{\AA}$ & (geometry filter) \\
\hline

Cation--$\pi$ interaction
& Cation--ring centroid distance
& $3.0\,\text{\AA}<d\leq 6.0\,\text{\AA}$
& $
s(d)=
\begin{cases}
\max\!\left(10^{-6},\,1-\dfrac{d-3.0\,\text{\AA}}{6.0\,\text{\AA}-3.0\,\text{\AA}}\right), & 3.0\,\text{\AA}<d\leq 6.0\,\text{\AA} \\
10^{-6}, & \text{otherwise}
\end{cases}
$ \\
\hline

Hydrophobic contact
& Distance between apolar atoms
& $3.0\,\text{\AA}<d\leq 5.0\,\text{\AA}$
& $
s(d)=
\begin{cases}
\max\!\left(10^{-6},\,1-\dfrac{d-3.0\,\text{\AA}}{5.0\,\text{\AA}-3.0\,\text{\AA}}\right), & 3.0\,\text{\AA}<d\leq 5.0\,\text{\AA} \\
10^{-6}, & \text{otherwise}
\end{cases}
$ \\
\hline

Van der Waals forces
& Distance between non-hydrogen atoms
& $d_0<d\leq 4.5\,\text{\AA}$, where $d_0=R_{\mathrm{vdW}}(A){+}R_{\mathrm{vdW}}(B)$
& $
s(d)=
\begin{cases}
\max\!\left(10^{-6},\,1-\dfrac{d-d_0}{4.5\,\text{\AA}-d_0}\right), & d_0<d\leq 4.5\,\text{\AA} \\
10^{-6}, & \text{otherwise}
\end{cases}
$ \\
\Xhline{0.8pt}
\end{tabular}}
\end{table*}

\section{Baselines}\label{baseline_detail}

We compare the following eight representative baselines:

\begin{enumerate}
    \item MolTrans~\cite{huang2021moltrans}, which tokenizes protein sequences and ligands into substructures, encodes them with transformer-based encoders, and applies an interaction module to model cross-modal binding patterns.
    
    \item TransformerCPI~\cite{chen2020transformercpi}, which represents proteins and compounds as sequences and employs transformer architectures to model their interactions for compound--protein interaction prediction.
    
    \item HyperAttentionDTI~\cite{zhao2022hyperattentiondti}, which introduces a hyper-attention mechanism to capture fine-grained interactions between protein residues and ligand atoms for drug--target interaction prediction.
    
    \item PerceiverCPI~\cite{nguyen2023perceiver}, which adopts a Perceiver-based architecture to encode long protein and ligand sequences and model compound--protein interactions through latent cross-attention.
    
    \item CAT-DTI~\cite{zeng2024cat}, which uses cross-attention mechanisms to jointly model protein and ligand representations and improve interaction prediction.
    
    \item DrugBAN~\cite{bai2023interpretable}, which encodes ligands with a graph neural network and proteins with a 1D CNN, and employs a bilinear attention network~\cite{kim2018bilinear} to capture fine-grained residue--atom interactions before final prediction.
    
    \item GraphBAN~\cite{hadipour2025graphban}, which extends bilinear attention-based interaction modeling with graph neural representations for ligands and protein encoders for interaction prediction.
    
    \item FusionDTI~\cite{meng-etal-2025-fusiondti}, which uses token-level fusion to model residue--atom interactions between protein and ligand representations for drug--target interaction prediction.
\end{enumerate}

\clearpage
\section{Ligand Similarity-controlled Dataset Splits}
\label{app:ligand_similarity_results}

\begin{figure*}[t]
    \centering
    \includegraphics[width=\textwidth]{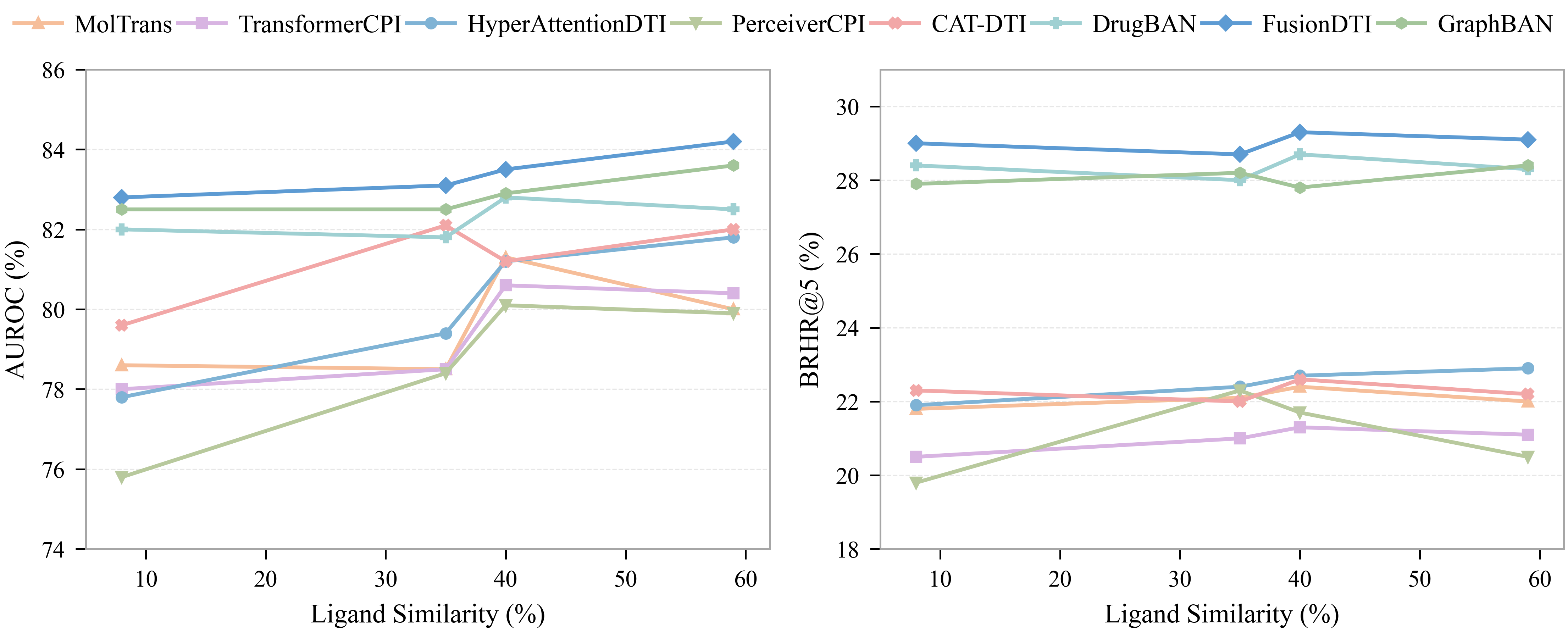}
    \caption{
    \textbf{Performance across four ligand similarity-controlled OOD datasets.}
    The four datasets have maximum train--test ligand similarities of 8\%, 35\%, 40\%, and 59\%, with a lower similarity indicating a stronger ligand-side distribution shift.
    \textbf{a.} Binary binding prediction performance measured by AUROC on the four datasets.
    \textbf{b.} Binding-site localization performance measured by BRHR@5 on the four datasets.
    }
    \label{fig:ligand_similarity_results}
\end{figure*}

For the OOD setting, we also construct ligand similarity-controlled evaluation scenarios to assess whether chemical dissimilarity between training and test ligands affects model generalization. Ligand chemical similarity is quantified using Tanimoto similarity over extended connectivity fingerprints (ECFP)~\cite{rogers2010extended}, yielding four splits with maximum train--test ligand similarities of 8\%, 35\%, 40\%, and 59\%. To isolate the effect of ligand-side distribution shift, protein similarity is fixed at 33\% across all splits. Each dataset is divided into training, validation, and test subsets with a ratio of 8:1:1. As shown in Fig.~\ref{fig:ligand_similarity_results}a, binary binding prediction remains relatively stable across the four ligand similarity-controlled splits. FusionDTI shows the strongest performance, increasing only slightly from 82.8\% AUROC at 8\% ligand similarity to 84.2\% at 59\%. GraphBAN similarly varies within a narrow range, from 82.5\% to 83.6\%, while DrugBAN remains stable between 81.8\% and 82.8\%. Other baselines also show only modest variation, for example CAT-DTI ranges from 79.6\% to 82.1\%, HyperAttentionDTI from 77.8\% to 81.8\%, and MolTrans from 78.5\% to 81.3\%. Overall, these results indicate that ligand similarity has a comparatively limited effect on binary binding prediction under the current InteractBind setting. Binding-site localization shows an even smaller change across ligand similarity-controlled splits (Fig.~\ref{fig:ligand_similarity_results}b). FusionDTI remains the strongest model, with BRHR@5 varying only slightly from 28.7\% to 29.3\%, while DrugBAN ranges from 28.0\% to 28.7\% and GraphBAN from 27.8\% to 28.4\%. The remaining baselines also fluctuate within narrow ranges, including MolTrans from 21.8\% to 22.4\%, HyperAttentionDTI from 21.9\% to 22.9\%, and CAT-DTI from 22.0\% to 22.6\%. Although PerceiverCPI shows somewhat less consistent behavior, its BRHR@5 still remains within a limited range of 19.8\% to 22.3\%. Compared with the protein similarity-controlled splits, these changes are substantially smaller, indicating that protein-side similarity exerts a stronger influence on model inference than ligand-side chemical similarity. One possible explanation is that binding-site localization is evaluated on the protein axis, making protein sequence diversity more directly related to the difficulty of identifying ligand-binding residues. These results support the use of protein similarity-controlled splits as the primary OOD evaluation setting, while ligand-controlled splits provide a complementary assessment of chemical generalization.

\clearpage

\section{Licenses and Terms of Use for Existing Resources}
\label{licenses}

In the following, we summarize the licenses and terms of use of the main existing resources used in this work.

\begin{itemize}
    \item \textbf{Protein Data Bank~\cite{burley2017protein}} Source: wwPDB / RCSB PDB archive. License: CC0 1.0 Universal (public domain dedication). We use experimentally resolved protein--ligand complex structures from the PDB for dataset construction, and attribute original depositors where appropriate.
    
    \item \textbf{PLIP~\cite{schake2025plip}} Source: \texttt{pharmai/plip}. License: GNU GPL v2.0. We use PLIP to identify hydrogen bonds, salt bridges, $\pi$--$\pi$ stacking, cation--$\pi$ interactions, and hydrophobic contacts during interaction annotation.
    
    \item \textbf{GetContacts~\cite{getcontacts}} Source: \texttt{getcontacts/getcontacts}. License: Apache License 2.0. We use GetContacts to compute van der Waals interactions not covered by PLIP.
    
    \item \textbf{AutoDock Vina~\cite{trott2010autodock}} Source: official AutoDock Vina release / \texttt{ccsb-scripps/AutoDock-Vina}. License: Apache License 2.0. We use AutoDock Vina for docking-based affinity estimation and negative-sample construction.
\end{itemize}

\clearpage


\newpage
\section*{NeurIPS Paper Checklist}

\begin{enumerate}

\item {\bf Claims}
    \item[] Question: Do the main claims made in the abstract and introduction accurately reflect the paper's contributions and scope?
    \item[] Answer: \answerYes{} 
    \item[] Justification: The abstract and introduction accurately state the paper's contributions: InteractBind is introduced as a large-scale protein--ligand dataset and benchmark for fine-grained binding-site evaluation, with residue--atom interaction maps covering six major non-covalent interaction types. The claims are supported by the dataset construction procedure, the benchmark protocol, and empirical evaluation of eight sequence-based and interaction-aware models.

\item {\bf Limitations}
    \item[] Question: Does the paper discuss the limitations of the work performed by the authors?
    \item[] Answer: \answerYes{} 
    \item[] Justification: The paper includes a limitations discussion. In particular, it states that the current benchmark focuses on sequence-level evaluation and does not yet include structure-level benchmarking, because three-dimensional protein--ligand evaluation involves pose ambiguity and conformational flexibility. This clarifies the scope of the current contribution and identifies structure-level evaluation as future work.

\item {\bf Theory assumptions and proofs}
    \item[] Question: For each theoretical result, does the paper provide the full set of assumptions and a complete (and correct) proof?
    \item[] Answer: \answerNA{} 
    \item[] Justification: The paper does not present theoretical results, theorems, or formal proofs. Its contributions are a dataset, benchmark design, evaluation protocol, and empirical analysis.

    \item {\bf Experimental result reproducibility}
    \item[] Question: Does the paper fully disclose all the information needed to reproduce the main experimental results of the paper to the extent that it affects the main claims and/or conclusions of the paper (regardless of whether the code and data are provided or not)?
    \item[] Answer: \answerYes{} 
    \item[] Justification: The paper describes the dataset construction pipeline, interaction annotation procedure, positive and negative sample construction, ID and OOD split protocols, baseline models, and evaluation metrics. These details provide a clear basis for reproducing or verifying the main experimental findings. Additional implementation details and release instructions are provided in the appendix and accompanying release package.

\item {\bf Open access to data and code}
    \item[] Question: Does the paper provide open access to the data and code, with sufficient instructions to faithfully reproduce the main experimental results, as described in supplemental material?
    \item[] Answer: \answerYes{} 
    \item[] Justification: The submission provides access to both the InteractBind dataset and the code used for benchmark evaluation, and reproduction of the main experimental results.

\item {\bf Experimental setting/details}
    \item[] Question: Does the paper specify all the training and test details (e.g., data splits, hyperparameters, how they were chosen, type of optimizer) necessary to understand the results?
    \item[] Answer: \answerYes{} 
    \item[] Justification: The paper specifies the main experimental settings, including the affinity-aware in-distribution split, protein similarity-controlled OOD splits, ligand similarity-controlled analysis, baseline models, evaluation metrics, and the reporting of mean and standard deviation. Additional baseline descriptions and dataset statistics are provided in the appendix.

\item {\bf Experiment statistical significance}
    \item[] Question: Does the paper report error bars suitably and correctly defined or other appropriate information about the statistical significance of the experiments?
    \item[] Answer: \answerYes{} 
    \item[] Justification: The main results are reported as mean $\pm$ standard deviation. In-distribution results are reported across 5-fold cross-validation, while OOD results are reported over five runs with different random seeds on predefined train/validation/test splits. The reported standard deviations capture variability across folds or random seeds, depending on the experiment.

\item {\bf Experiments compute resources}
    \item[] Question: For each experiment, does the paper provide sufficient information on the computer resources (type of compute workers, memory, time of execution) needed to reproduce the experiments?
    \item[] Answer: \answerYes{} 
    \item[] Justification: The appendix specifies the hardware used for the experiments.
    
\item {\bf Code of ethics}
    \item[] Question: Does the research conducted in the paper conform, in every respect, with the NeurIPS Code of Ethics \url{https://neurips.cc/public/EthicsGuidelines}?
    \item[] Answer:\answerYes{}
    \item[] Justification: The authors have reviewed the NeurIPS Code of Ethics. The work is intended to support transparent and responsible evaluation of protein--ligand modelling systems. It uses existing structural resources and computational annotation pipelines, and does not involve human subjects, private personal data, or high-risk generative models.

\item {\bf Broader impacts}
    \item[] Question: Does the paper discuss both potential positive societal impacts and negative societal impacts of the work performed?
    \item[] Answer: \answerYes{} 
    \item[] Justification: The paper discusses broader impacts, including the potential positive impact of promoting more interpretable and physically grounded protein--ligand modelling. It also notes possible negative consequences if inferred binding sites or interaction patterns are used without experimental validation, as incorrect mechanistic interpretation may lead to misleading biological conclusions or suboptimal compound prioritisation.
    
\item {\bf Safeguards}
    \item[] Question: Does the paper describe safeguards that have been put in place for responsible release of data or models that have a high risk for misuse (e.g., pre-trained language models, image generators, or scraped datasets)?
    \item[] Answer: \answerNA{} 
    \item[] Justification: The work does not release a high-risk generative model, a large language model, or an Internet-scraped dataset of the type highlighted in the checklist. The released asset is a protein--ligand evaluation dataset and benchmark derived from existing structural resources.

\item {\bf Licenses for existing assets}
    \item[] Question: Are the creators or original owners of assets (e.g., code, data, models), used in the paper, properly credited and are the license and terms of use explicitly mentioned and properly respected?
    \item[] Answer: \answerYes{} 
    \item[] Justification: The paper cites the existing resources and tools used in the study, including the Protein Data Bank, PLIP, GetContacts, and AutoDock Vina. The appendix reports their licences and terms of use where available, and describes how each resource is used in the dataset construction and annotation pipeline.

\item {\bf New assets}
    \item[] Question: Are new assets introduced in the paper well documented and is the documentation provided alongside the assets?
    \item[] Answer: \answerYes{} 
    \item[] Justification: The paper introduces InteractBind as a new dataset and benchmark. It documents the dataset construction process, sample contents, annotation types, split protocols, evaluation tasks, and metrics in the main paper and appendix. The dataset release includes the information needed to understand and use the benchmark. 

\item {\bf Crowdsourcing and research with human subjects}
    \item[] Question: For crowdsourcing experiments and research with human subjects, does the paper include the full text of instructions given to participants and screenshots, if applicable, as well as details about compensation (if any)? 
    \item[] Answer: \answerNA{} 
    \item[] Justification: The work does not involve crowdsourcing experiments or research with human subjects. The dataset is constructed from existing protein--ligand structural resources and computational annotation pipelines.

\item {\bf Institutional review board (IRB) approvals or equivalent for research with human subjects}
    \item[] Question: Does the paper describe potential risks incurred by study participants, whether such risks were disclosed to the subjects, and whether Institutional Review Board (IRB) approvals (or an equivalent approval/review based on the requirements of your country or institution) were obtained?
    \item[] Answer: \answerNA{} 
    \item[] Justification: The work does not involve human participants, crowdsourcing, or collection of human-subject data requiring IRB review.

\item {\bf Declaration of LLM usage}
    \item[] Question: Does the paper describe the usage of LLMs if it is an important, original, or non-standard component of the core methods in this research? Note that if the LLM is used only for writing, editing, or formatting purposes and does \emph{not} impact the core methodology, scientific rigor, or originality of the research, declaration is not required.
    \item[] Answer: \answerNA{} 
    \item[] Justification: LLMs are not an important, original, or non-standard component of the dataset construction, benchmark design, or experimental methodology in this work.

\end{enumerate}

\end{document}